\documentclass[conference]{IEEEtran}
\IEEEoverridecommandlockouts
\usepackage{cite}
\usepackage{amsmath,amssymb,amsfonts}
\usepackage{algorithmic}
\usepackage{algorithm}
\usepackage{graphicx}
\usepackage{textcomp}
\usepackage{xcolor}
\usepackage{url}
\usepackage{amsfonts}
\usepackage{bm}
\usepackage{multirow}

\newcommand{\nop}[1]{}

\def\BibTeX{{\rm B\kern-.05em{\sc i\kern-.025em b}\kern-.08em
    T\kern-.1667em\lower.7ex\hbox{E}\kern-.125emX}}
\begin{document}

\title{Interpretable Distance Metric Learning for Handwritten Chinese Character Recognition}

\author{
\IEEEauthorblockN{
Boxiang Dong\IEEEauthorrefmark{1}, Aparna S. Varde\IEEEauthorrefmark{1}, Danilo Stevanovic\IEEEauthorrefmark{1}
Jiayin Wang\IEEEauthorrefmark{1},
Liang Zhao\IEEEauthorrefmark{2}
}
\IEEEauthorblockA{\IEEEauthorrefmark{1}\{dongb, vardea, stevanovicd1, jiayin.wang\}@montclair.edu}
Montclair State University, Montclair, New Jersey 07043\\
\IEEEauthorblockA{\IEEEauthorrefmark{2}
liangzhao@dlut.edu.cn \\
Dalian University of Technology, Dalian, China 116024
}
}

\maketitle

\begin{abstract}
% importance
Handwriting recognition is of crucial importance to both Human Computer Interaction (HCI) and paperwork digitization.
% difficulty
In the general field of Optical Character Recognition (OCR), handwritten  Chinese  character  recognition faces tremendous challenges due to the enormously large character sets and the amazing diversity of writing styles.
% distance metric learning
Learning an appropriate distance metric to measure the difference between data inputs is the foundation of accurate handwritten character recognition.
Existing distance metric learning approaches either produce unacceptable error rates, or provide little interpretability in the results.
% component, interpretable
In this paper, we propose an interpretable distance metric learning approach for handwritten  Chinese  character  recognition. The learned metric is a linear combination of intelligible base metrics, and thus provides meaningful insights to ordinary users.
% experiment
Our experimental results on a benchmark dataset demonstrate the superior efficiency, accuracy and interpretability of our proposed approach.
\end{abstract}


%AV: Boxiang, usually, I try not to repeat keywords that are already in the title, so I am trying to modify these. Somehow, I want to emphasize the fact that Chinese text has a pictorial script (that is why we considered approaches used for graphical data). I also added Big Data since the characters are many in number & variety. I think it is good to put HCI & OCR too. Feel free to alter these keywords as needed. 

\section{Introduction}
\label{sc:intro}
% HCCR
    % importance and application 
    % AV: Excellent, this motivation is so appealing!
Handwriting recognition is gaining increasing importance given the prevalence of mobile devices and tablets. A majority of people still prefer handwritten input over keyboard entry, especially when taking notes in a classroom or meeting, and annotating a digital document.
The need for an accurate and reliable handwriting recognition solution is even stronger among Chinese users, given the fact that it is extremely time-consuming to enter a Chinese character via keyboard \cite{apple}. In particular,
users have to type in the pronunciation (i.e., Pinyin) of the desired Chinese character first, and then choose the target from a list of candidates. To make it even more difficult, the pronunciation of a single Chinese character usually consists of at least 4 English characters. 
On the other hand, handwriting recognition is the fundamental technology of Optical Character Recognition, which is widely applied in handwritten check clearance and judicial paperwork digitization. 

% distance metric learning
Pairwise distance metric, a function that measures the dissimilarity between a pair of data inputs, plays crucial role in handwritten character recognition. Ideally, it helps to identify the handwriting inputs that correspond to the same character. It is obvious that the performance of a handwritten Chinese character recognition (HCCR) system heavily depends on the quality of the underlying distance metric.
Distance metric learning (DML) aims at automatically learning an appropriate distance (or similarity) measure from labeled samples \cite{chopra2005learning,bellet2013survey}. Recent results \cite{goldberger2004nips} reveal that even a simple linear transformation of the input features can significantly improve the classification accuracy. Therefore, DML provides a natural solution to determine the distance metric for handwritten Chinese character recognition.
%AV: Boxiang, sorry I deleted a reference from the bib by mistake and then tried to find it based on the contents herewith and added that. It is the NIPS paper by Goldberg et al. on Neighborbhood. 

% not accurate, uninterpretable: feature + model
% AV: Nice! I did not know that this was the only work on DML in Chinese char recognition (I just put a disclaimer in brackets if that is okay with you).
Surprisingly, \cite{yin2009handwritten} is the only work on distance metric learning for handwritten Chinese character recognition (based on our literature search). However, the application of the distance metric learning there is only limited to text line segmentation. 
Handwritten Chinese character recognition and distance metric learning do face their own challenges respectively. 
In terms of handwritten Chinese character recognition, the challenges arise from the enormous character set and the diversity of writing styles. Unlike alphabet-based writing, which typically comprises the order of 100 symbols, there are 27,533 entries in Chinese National Standard GB18030-2005. Moreover, the divergence of writing styles among different writers and in different geographic areas aggravates the confusion between different characters \cite{wang2011handwritten}.
These difficulties lead to unsatisfactory performance in handwriting recognition. For example, \cite{su2009off} can only provide 39.37\% recognition accuracy on a dataset with 186,444 characters. Later works \cite{wang2009integrating,li2010bayesian} improve the accuracy up to 78.44\% and 73.87\% respectively.
Regarding distance metric learning, most works focus on learning a Mahalanobis distance metric. Even though it is equivalent to computing the Euclidean distance after a linear projection of the data, it provides limited interpretability. In particular, the learned transformation matrix cannot explain the relative importance of the features in the distance metric. Lacking interpretability prohibits further analysis in the case of misclassification and undermines user confidence.

% objective
In this paper, our objective is to put forth an interpretable distance metric learning approach for accurate offline handwritten Chinese character recognition. By offline, we mean that the focus is on recognizing characters already written on paper earlier. The input is in the form of a scanned image of the paper document. 
Compared with online recognition, offline character recognition is more challenging in that it does not have the trace of the writer's pen as well as the order of writing in the input. It has been shown \cite{nishimura2003offline} that such pen dynamics information can help to obtain better recognition accuracy than static scanned images alone. 

% transition to components, and select components
To provide interpretability in the learned distance metric, we firstly define a set of base distance metrics, which we call \emph{components} in the rest of the paper. These components quantify the dissimilarity of two handwritten characters in a simple manner. They can be provided by domain experts in fields such as Linguistics, or can be proposed based on a preliminary analysis of the data. 

The components used in our experiments include the difference of the length of the longest horizontal stroke, the longest vertical stroke, etc.
Given these components, we propose an ensemble learning strategy to linearly combine these basic metrics into a strong  distance metric, so as to guarantee accurate handwritten character recognition.

To the best of our knowledge, ours is among the first works to learn an interpretable distance metric for handwritten Chinese character recognition.
In particular, this work makes the following contributions.

% contribution
% AV: Boxiang, we will have to change the name of this algorithm. This looks quite different from the LearnMet approach that I had proposed in my earlier work. In any case we cannot repeat names, since LearnMet has already been published in a journal. For now, I am calling this MetChar, since it is to obtain a metric for characters. 

\begin{itemize}
    \item We design a new algorithm named {\em MetChar} that optimizes the weight assignment for a given set of basic components so as to obtain a distance metric, which is later used by the clustering algorithm to classify the input handwritten characters. {\em MetChar} follows a style analogous to the gradient descend optimizer \cite{kingma2014adam}, but it copes with the fact that the error rate is not differentiable. A good property of {\em MetChar} is that it is compatible with a wide range of clustering algorithms.
    \item We propose an approach, namely {\em HybridSelection}, that chooses the combination of basic components from a large candidate pool, and feeds them to {\em MetChar} for optimization. The {\em HybridSelection} algorithm trims off the components that do not meet the quality requirement, and fully takes advantage of the remaining ones. It reaches a balance between efficiency and accuracy.
    \item We run a set of experiments on a benchmark dataset. The results demonstrate the superiority of {\em HybridSelection}. It produces the highest recognition accuracy in an affordable time. Besides, the learned distance metric is interpretable for ordinary users.
\end{itemize}

The rest of the paper is organized as follows. 
Section \ref{sc:pre} reviews the preliminaries. Section \ref{sc:method} presents our distance metric learning approaches. Section \ref{sc:exp} discusses the experimental results. Section \ref{sc:related} introduces the related work. Finally, Section \ref{sc:conclu} concludes the paper.

%1.2 page
\section{Preliminaries}
\label{sc:pre}
% individual components 
\subsection{Individual Distance Components}
\label{sc:components}
\begin{figure}[!hbtp]
        \centering
        \includegraphics[width=0.48\textwidth]{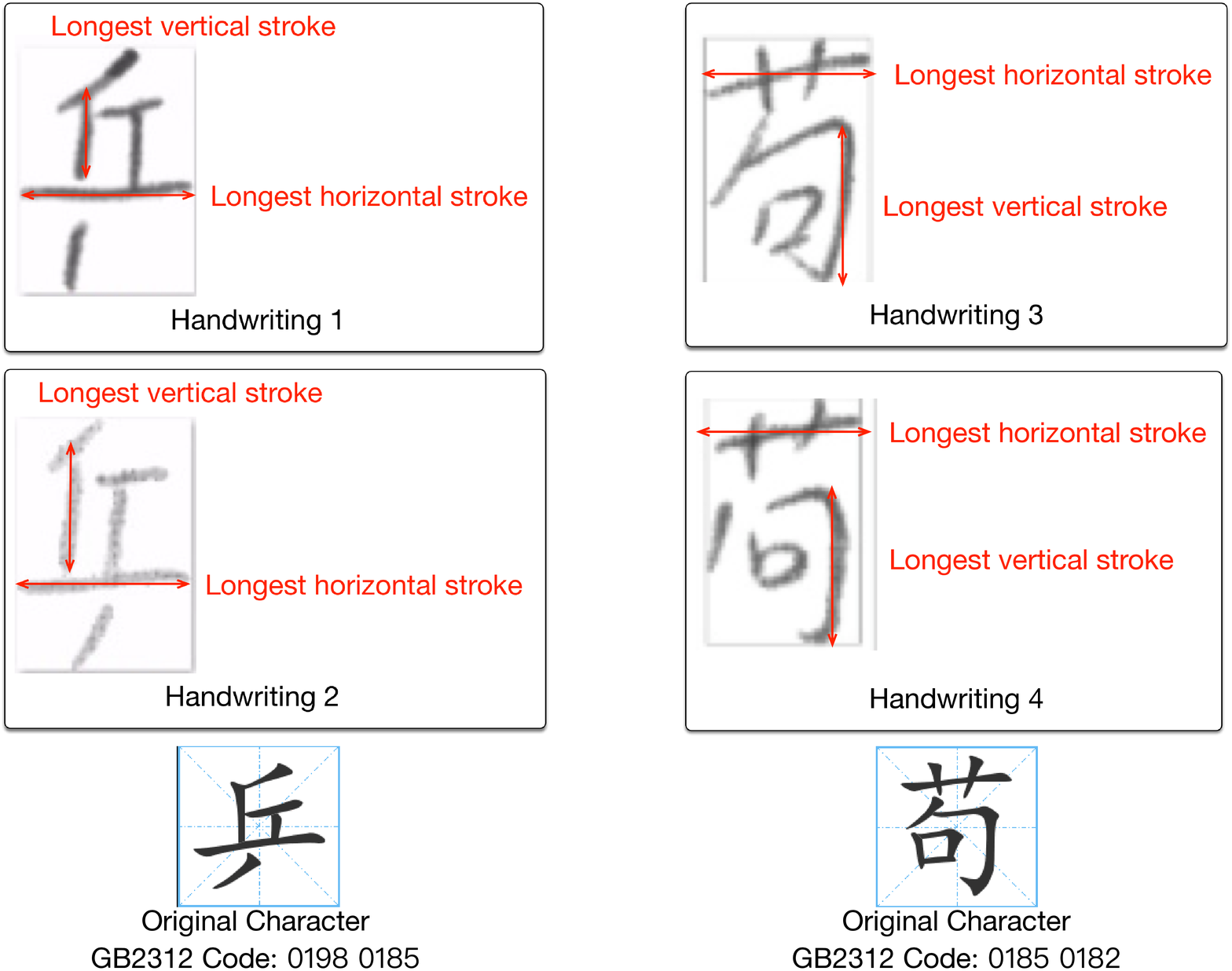}
        \caption{An example of individual distance components in handwritten Chinese character recognition}
        \label{fig:components}
    \end{figure}

    % CV, regions
    In computer vision and data management, it is well known that a small fraction of the regions in a figure can hold critical information for object recognition and classification \cite{wang2018ictai, varde2007icde, varde2006sigmod, vardedistance, varde2008component}. In addition, certain statistical studies on the figure depict the contextual behavior in the image in a succinct manner \cite{carbonetto2004statistical}.
    % Handwritting 
    In the field of handwritten Chinese character recognition, we have similar observations. Take the handwritten characters in Fig. \ref{fig:components} as an example. The length and position of the longest horizontal and vertical strokes in a handwritten character can serve as effective and convenient features for the purpose of recognition. Next, the individual distance components are defined by applying simple operators on these intelligible features, such as Euclidean distance and Manhattan distance.
    % Example
    % Advantage
    As opposed to sophisticated feature engineering techniques of crossings and celled projections \cite{hossain2012rapid,shen2012positive}, the components used in this paper are basic ones that incorporate fundamental human intelligence, and are more easily interpretable.

% distance metric learning
\subsection{Distance Metric Learning}
    In this paper, we aim at proposing an interpretable distance metric learning approach for handwritten Chinese character recognition. To provide interpretability, our objective is to learn a distance metric that is a linear combination of the intelligible individual components. In particular, given a set of individual components (metrics) $\bm{d} = \{d_1, \dots, d_p\}$, where $d_i(\bm{x}, \bm{x}')$ is per the discussion in Section \ref{sc:components}, the target distance metric is 
    \begin{equation}
    \label{eq:target}
            D(\bm{x}, \bm{x}')=\sum_{i=1}^{p} w_i d_i(\bm{x}, \bm{x}'),
    \end{equation}
    where $w_i \geq 0$ for every $1\leq i \leq p$, $\bm{x}$ and $\bm{x}'$ are a pair of handwriting characters. Due to the linear combination in Equation (\ref{eq:target}), the weight associated with each basic component signifies its importance in handwritten character recognition, so as to provide interpretability in the decision/classification result. It can be clearly seen that our objective is indeed a bagging algorithm to combine multiple basic but intelligible individual components, so as to formulate an accurate and interpretable distance metric. 
 % 0.8
\section{Distance Metric Learning Approach}
\label{sc:method}
In our approach, we first propose an algorithm named {\em MetChar} that optimizes the weight assignment for a fixed set of components. Next, we propose an algorithm named {\em HybridSelection} to select components as the input to {\em MetChar}.

\subsection{The {\em MetChar} Algorithm}
    % general intro: clustering, fixed components
    {\em MetChar} recognizes/classifies the input handwritten characters based on clustering. It is worth noting that {\em MetChar} is compatible with any clustering algorithm. Given a set of individual components $d_1, \dots, d_q \subseteq \bm{d}$, where $\bm{d}$ is the complete set of available components, {\em MetChar} produces the component weights $w_1, \dots, w_q$ so that the clustering algorithm based on the distance metric $D(\bm{x}, \bm{x}')=\sum_{i=1}^q w_id_i(\bm{x}, \bm{x}')$ assigns the handwriting samples of the same character into the same cluster, and those of different characters into different clusters. 
    % FP, TP 
    Before explaining the optimization procedure, we first present a few definitions that are needed later. Given a pair of handwriting samples $\bm{x}$ and $\bm{x}'$, let their corresponding characters be $y$ and $y'$ respectively. Let $c$ and $c'$ denote the clusters to which they are assigned. We then have the following definitions based on the relationship between the characters and the clusters.
    \begin{itemize}
        \item $\big(\bm{x}, \bm{x}'\big)$ is a true positive (TP) if $y=y'$ and $c=c'$.
        \item $\big(\bm{x}, \bm{x}'\big)$ is a true negative (TN) if $y\neq y'$ and $c\neq c'$.
        \item $\big(\bm{x}, \bm{x}'\big)$ is a false positive (FP) if $y\neq y'$ but $c=c'$.
        \item $\big(\bm{x}, \bm{x}'\big)$ is a false negative (FN) if $y=y'$ but $c\neq c'$.
    \end{itemize}
    It is obvious that TPs and TNs are the pairs that are correctly recognized. Hence we define the accuracy as $acc = \frac{TP+TN}{TP+TN+FP+FN}$.
    
    % Intuition to reduce error rate
    In order to improve the recognition accuracy, the intuition is to increase the pairwise distance for the FPs, while reducing the distance for the FNs. 
    At the $t$-th round of the learning stage, let $D^{(t)}$ be the learned distance metric, and $w^{(t)}_i$ denote the weight associated with the individual component $d_i$.
    For any $d_i$, we can calculate two values: $\alpha_i^{(t)}$ which denotes the total distance on $d_i$ for the FPs, and $\beta_i^{(t)}$ which denotes the total distance on $d_i$ for the FNs. In particular, we have 
    \begin{equation}
    \label{eq:a}
        \alpha_i^{(t)} = \sum_{(\bm{x}, \bm{x}')\in FP} w^{(t)}_i d_i\big(\bm{x}, \bm{x}'\big),
    \end{equation}
    and 
    \begin{equation}
    \label{eq:b}
        \beta_i^{(t)} = \sum_{(\bm{x}, \bm{x}')\in FN} w^{(t)}_i d_i\big(\bm{x}, \bm{x}'\big).
    \end{equation}
    In the next round, i.e., the $(t+1)$-th round, we update the weight associated with $d_i$ by:
    \begin{equation}
    \label{eq:update}
        w^{(t+1)}_i = max(0, w^{(t)}_i+\epsilon(\alpha_i^{(t)} - \beta_i^{(t)})),
    \end{equation}
    where $\epsilon$ is a given learning rate. We repeatedly update the weights $w_1, \dots, w_q$ for a certain number of iterations. Algorithm \ref{alg:metchar} displays the pseudocode for {\em MetChar}.
    
    % Pseudocode
\begin{algorithm} % enter the algorithm environment
\caption{{\em MetChar}} % give the algorithm a caption
\label{alg:metchar} % and a label for \ref{} commands later in the document
\begin{algorithmic}[1] % enter the algorithmic environment
    \REQUIRE A set of individual components $d_1, \dots, d_q$, the training set $\{(\bm{x}, y)\}$, the learning rate $\epsilon$, the number of iterations $T$, the number of unique characters $k$
    \ENSURE The distance metric $D=\sum_{i=1}^q w_i d_i(\bm{x}, \bm{x}')$
    \STATE Randomly assign initial weights $w_1^{(1)}, \dots, w_q^{(1)}$
    \FOR{$t=1$ to $T$}
        \STATE{Let $D^{(t)} = \sum_{i=1}^q w_i^{(t)} d_i(\bm{x}, \bm{x}')$}
        \STATE{Apply the clustering algorithm with $D^{(t)}$ to get $k$ clusters}
        \STATE{Calculate the accuracy $acc^{(t)}$}
        \FOR{$i=1$ to $q$}
            \STATE{Update $w^{(t+1)}_i$ according to Equation (\ref{eq:update})}
        \ENDFOR
    \ENDFOR
    
    \STATE{Let $t^*$ be the round that produces the highest $acc^{(t^*)}$}
    \RETURN{$D^{(t^*)}$}
\end{algorithmic}
\end{algorithm}

{\em MetChar} follows a style similar to the classical gradient descent algorithm \cite{kingma2014adam}, which is widely adapted in deep learning. In particular, they both start with a random initialization of the weights, and then gradually optimize the weights throughout the iterations. However, since the underlying recognition algorithm is clustering, the loss value or error rate is not differentiable over the weights. In other words, it is impossible to calculate the derivatives of the loss over the weights. Therefore, unlike the original gradient descent algorithm, {\em MetChar} refines the weights by taking the FPs and FNs into consideration (Equation (\ref{eq:a} - \ref{eq:update})).

\subsection{Component Selection Algorithms}
% LearnMet fully use components
The {\em MetChar} algorithm takes a set of individual components as the input, and by default utilizes all of them to learn a distance metric for handwritten Chinese character recognition. However, in practice, it is not necessary to exploit all the components, especially if a large candidate pool is available. 
% not necessary, overfitting
Introducing redundant components demands longer optimization time of {\em MetChar}, while bringing minimal accuracy benefits. Moreover, since the number of parameters to be optimized is linear to the number of components, redundant components could make the learned distance metric overfit the training data, and lead to a suboptimal solution \cite{qian2014distance}. 
% Need for component selection
Therefore, it is imperative to have an approach for selecting a subset of components from the candidate pool which {\em MetChar} can employ to deliver satisfactory recognition accuracy.

A naive solution is to enumerate all possible combinations of components and feed them to {\em MetChar}. However, this incurs significant computational overhead. The search complexity is exponential to the number of components, i.e. $O(2^p)$. This is overwhelming when $p$ is a large number. 
To reach a balance between efficiency and accuracy, we propose {\em HybridSelection} that firstly eliminates the least promising candidate components, based on a given error threshold, and then examines all the combinations of the remaining components.

\begin{algorithm} % enter the algorithm environment
\caption{{\em HybridSelection}} % give the algorithm a caption
\label{alg:hybrid} % and a label for \ref{} commands later in the document
\begin{algorithmic}[1] % enter the algorithmic environment
    \REQUIRE A complete set of individual components $\bm{d} = \{d_1, \dots, d_p\}$, an error threshold $\theta$
    \ENSURE The distance metric $D$
    
    \FOR{$j=1$ to $p$}
        \STATE{Run {\em MetChar} (Algorithm \ref{alg:metchar}) with $d_i$ only}
        \STATE{Evaluate the accuracy $acc_i$}
        \STATE{Prune $d_i$ from $\bm{d}$ if $acc_i < \theta$}
    \ENDFOR
    \FOR{$j=2$ to $p$}
        \FOR{each combination $\bm{d}'$ of $j$ components in $\bm{d}$}
            \STATE{Call {\em MetChar} (Algorithm \ref{alg:metchar}) on $\bm{d}'$}
        \ENDFOR
    \ENDFOR
    \RETURN{the distance metric with the highest accuracy}
\end{algorithmic}
\end{algorithm}

Algorithm \ref{alg:hybrid} shows the pseudocode. From Line 1 - 5, {\em HybridSelection} evaluates the quality of each individual component. If $d_i$ does not meet the quality requirement, i.e., $acc_i < \theta$ (error threshold), it is eliminated from the component pool. From Line 6 to 10, {\em HybridSelection} tries every combination of the remaining components, and finds the one with the highest accuracy. Let $s$ denote the number of remaining components after the first loop, the complexity is $O(p+2^s)$, where $s<p$. Therefore, the complexity is lower than that of {\em ExhaustiveSelection}. Meanwhile, we can adjust the error threshold $\theta$ to alternate the balance between efficiency and accuracy. A larger $\theta$ results in a smaller $s$, and thus better efficiency but potentially lower accuracy. 
Our experimental results in Section \ref{sc:exp} demonstrate that {\em HybridSelection} delivers satisfactory accuracy. 
%It is obviously higher than that of the {\em GreedySelection} approach. 
This is because the search space discarded by {\em HybridSelection} only includes the least promising combination of components, and thus induces little impact to the recognition accuracy. Given the same time constraints, the accuracy of {\em HybridSelection} can even be higher than that of the exhaustive search algorithm.
 % 2 
\begin{table}[!htbp]
    \centering
    \caption{Description of the CASIA-HWDB1.1 dataset}
    \begin{tabular}{|c|c|c|c|}
        \hline
        Dataset & \# of Writers & \# of Classes & \# of Sample Images \\\hline
        CASIA-HWDB1.1 & 300 & 3,755 & 1,121,749 \\\hline
    \end{tabular}
    \label{tb:data}
\end{table}

\begin{table*}[!hbtp]
    \centering
    \caption{Performance of {\em HybridSelection} with at least three components}
    \label{tb:hybrid}
    \begin{tabular}{|c|c|c|c|}
        \hline
        Components & Weights & Time (s) & Accuracy  \\\hline
        [$vlv\_md$, $dlv\_ed$, $hbv\_md$, $dfv\_md$] & [7.766, 5.803, 2.788, 3.197E-12] & 24.805 & 0.6650  \\\hline
        [$vlv\_md$, $dlv\_ed$, $hbv\_md$, $dfv\_ed$] & [6.152, 10.16, 1.267, 7.782] & 29.717 & 0.6802 \\\hline
        [$vlv\_md$, $dlv\_ed$, $hbv\_md$, $vfv\_md$] & [6.515, 9.432, 0.08981, 3.035] & 27.929 & 0.7218 \\\hline
        [$vlv\_md$, $dlv\_ed$, $dfv\_md$, $dfv\_ed$] & [1.623, 5.327, 0.2230, 4.277] & 31.191 & 0.7602 \\\hline
        [$vlv\_md$, $dlv\_ed$, $dfv\_md$, $vfv\_md$] & [7.116, 16.63, 4.711E-4, 5.391] & 26.015 & 0.6832  \\\hline
        [$vlv\_md$, $dlv\_ed$, $dfv\_ed$, $vfv\_md$] & [1.385, 10.33, 7.089, 0.8096] & 29.067 & 0.7993 \\\hline
        [$vlv\_md$, $hbv\_md$, $dfv\_md$, $dfv\_ed$] & [3.265, 4.753, 0.002662, 13.27] & 35.122 & 0.7960 \\\hline
        [$vlv\_md$, $hbv\_md$, $dfv\_md$, $vfv\_md$] & [6.675, 4.803, 1.493E-13, 4.174] & 58.505 & 0.3684 \\\hline
        [$vlv\_md$, $hbv\_md$, $dfv\_ed$, $vfv\_md$] & [3.999, 4.336, 13.61, 0.001724] & 35.008 & 0.6466 \\\hline
        [$vlv\_md$, $dfv\_md$, $dfv\_ed$, $vfv\_md$] & [5.235, 2.305, 13.31, 0.86907] & 39.629 & 0.7120 \\\hline
        [$dlv\_ed$, $hbv\_md$, $dfv\_md$, $dfv\_ed$] & [5.642, 0.7848, 4.593E-5, 4.542] & 34.291 & 0.6791 \\\hline
        [$dlv\_ed$, $hbv\_md$, $dfv\_md$, $vfv\_md$] & [14.54, 0.2431, 8.6984E-5, 4.491] & 31.894 & 0.6993 \\\hline
        [$dlv\_ed$, $hbv\_md$, $dfv\_ed$, $vfv\_md$] & [5.751, 0.1945, 4.172, 0.4177] & 36.026 & 0.6715 \\\hline
        [$dlv\_ed$, $dfv\_md$, $dfv\_ed$, $vfv\_md$] & [11.55, 0.012576, 8.236, 0.9782] & 33.655 &  0.6363 \\\hline
        [$hbv\_md$, $dfv\_md$, $dfv\_ed$, $vfv\_md$] & [4.438, 0.008846, 13.23, 0.002252] & 42.801 & 0.7562 \\\hline
        [$vlv\_md$, $dlv\_ed$, $hbv\_md$, $dfv\_md$, $dfv\_ed$] & [9.344, 12.51, 1.414, 0.03516, 9.257] & 36.644 & 0.7512 \\\hline
        [$vlv\_md$, $dlv\_ed$, $hbv\_md$, $dfv\_md$, $vfv\_md$] & [3.633, 20.90, 0.6605, 0.001985, 6.226] & 30.902 & 0.7191 \\\hline
        [$vlv\_md$, $dlv\_ed$, $hbv\_md$, $dfv\_ed$, $vfv\_md$] & [4.723, 8.905, 0.4436, 5.994, 0.5144] & 34.465 & 0.7279 \\\hline
        [$vlv\_md$, $dlv\_ed$, $dfv\_md$, $dfv\_ed$, $vfv\_md$] & [1.460, 12.54, 0.030117, 8.365, 0.9173] & 34.917 & 0.7089 \\\hline
        [$vlv\_md$, $hbv\_md$, $dfv\_md$, $dfv\_ed$, $vfv\_md$] & [5.694, 4.262, 0.001655, 12.48, 9.859E-4] & 42.443 & {\bf 0.8075} \\\hline
        [$dlv\_ed$, $hbv\_md$, $dfv\_md$, $dfv\_ed$, $vfv\_md$] & [10.70, 0.5038, 0.003774, 7.932, 0.5845] & 42.591 & 0.7729 \\\hline
        [$vlv\_md$, $dlv\_ed$, $hbv\_md$, $dfv\_md$, $dfv\_ed$, $vfv\_md$] &  [1.252, 0.5961, 19.19, 11.281, 4.382, 2.560] & 41.859 & 0.7227 \\\hline
    \end{tabular}
\end{table*}

\section{Experiments}
\label{sc:exp}

\subsection{Setup}
We implement the distance metric learning approaches in Java. We use k-means clustering in the {\em MetChar} algorithm. The source code is publicly available\footnote{\url{https://github.com/bxdong7/DML4HCCR}}. We execute all the experiments on a MacBook Pro with 3.1 GHz Intel i5 CPU and 16 GB RAM, running Mac OS X.

% data
\subsection{Dataset}
% name
We run experiments on the CASIA Offline Chinese Handwriting Database V1.1\footnote{\url{http://www.nlpr.ia.ac.cn/databases/handwriting/Home.html}}, namely CASIA-HWDB1.1. This is built by the National Laboratory of Pattern Recognition, Institute of Automation of Chinese Academy of Sciences.
% # of writers
The dataset is produced by 300 writers using Anoto pen on papers for obtaining offline images (in resolution of 300DPT).
The collected images are segmented and annotated at the character level.
The dataset includes 1,121,749 writing samples of 3,755 unique GB2312-80 level-1 Chinese characters. 
Table \ref{tb:data} presents the details of the CASIA-HWDB dataset. 

Every image in this dataset has its background labeled as 255 and its foreground pixels in 255 gray levels (0-254). In our experiments, we randomly choose 10 unique characters. The training and testing datasets include 30 non-overlapping sample images for each character. 

\subsection{Distance Components}
We preprocess each image by simply changing the foreground pixels to 1 and the background pixels to 0 so as to obtain the binary image. 
From each image, we extract the following features (see Fig. 1).
\begin{itemize}
    \item $hbv$: the horizontal bit vector, which denotes the number of $1$s in each horizontal line;
    \item $hfv$: the horizontal first foreground bit vector, which stores the location of the first $1$ in each horizontal line;
    \item $hlv$: the horizontal last foreground bit vector, which stores the location of the last $1$ in each horizontal line;
    \item $vfv$: the vertical first foreground bit vector, which stores the location of the first $1$ in each vertical line;
    \item $vlv$: the vertical last foreground bit vector, which stores the location of the last $1$ in each vertical line;
    \item $dfv$:  the diagonal first foreground bit vector, which stores the location of the first $1$ in each diagonal line; and
    \item $dlv$:  the diagonal last foreground bit vector, which stores the location of the last $1$ in each diagonal line.
\end{itemize}

Based on these features, we have the following basic distance metric components.
\begin{itemize}
    \item $hbv\_md$: the Manhattan distance between the $hbv$s of a pair of sample images;
    \item $hfv\_md$: the Manhattan distance between the $hfv$s of a pair of sample images;
    \item $vfv\_md$: the Manhattan distance between the $vfv$s of a pair of sample images;
    \item $vfv\_ed$: the Euclidean distance between the $vfv$s of a pair of sample images;
    \item $dfv\_md$: the Manhattan distance between the $dfv$s of a pair of sample images;
    \item $dfv\_ed$: the Euclidean distance between the $dfv$s of a pair of sample images;
    \item $hlv\_md$: the Manhattan distance between the $hlv$s of a pair of sample images;
    \item $vlv\_md$: the Manhattan distance between the $vlv$s of a pair of sample images;
    \item $vlv\_ed$: the Euclidean distance between the $vlv$s of a pair of sample images; and
    \item $dlv\_ed$: the Euclidean distance between the $dlv$s of a pair of sample images.
\end{itemize}

\subsection{Baselines}
In the experiment, we compare our component selection algorithm named {\em HybridSelection} with the following baselines:
\begin{itemize}
    \item {\bf ExhaustiveSelection} This method enumerates all possible combinations of individual distance components.
    \item {\bf GreedySelection} It firstly evaluates the recognition accuracy by using single component, and produces a canonical ordering of the individual components based on the accuracy. When generating a combination of $j>2$ components, it only considers the union of the $j$ best individual components.
\end{itemize}

\subsection{Performance of Evaluation}
Table \ref{tb:hybrid} displays the performance of {\em HybridSelection} with at least three components. 
In our experiments, we set $\theta=0.55$ based on grid search. This value of $\theta$ implies that those components producing a recognition accuracy lower than 0.55 are pruned out in Line 1 - 5 of Algorithm \ref{alg:hybrid}. This operation results in the six remaining components for consideration in Line 6 - 10 of Algorithm \ref{alg:hybrid}. 
From Table \ref{tb:hybrid}, we can see that {\em HybridSelection} completes examining a majority of the candidate combinations within 10 minutes. Moreover, the accuracy, 0.8075, i.e., $80.75 \%$ (as highlighted in the table) is acceptable for use in applications. 
%even higher than that of {\em ExhaustiveSelection}. This proves the superiority of {\em HybridSelection}. 

We also observe that the learned distance metric is interpretable as follows.  
Based on the associated weights, we can learn that among the 5 components whose combination provides the highest accuracy, $vlv\_md$, $hbv\_md$ and $dfv\_ed$ play the most important role. {\bf These components yield an accuracy of 80.7$\%$, i.e. $\sim$ 81$\%$} as observed from the experiments. It implies that the \emph{horizontal bit vector, vertical last foreground bit vector, and diagonal first foreground bit vector} are of crucial value for handwritten Chinese character recognition. These are highly meaningful insights gained from our study, with respect to metric interpretation. 

\begin{table}[!hbpt]
    \centering
    \caption{Comparison of {\em HybridSelection} with the baselines}
    \label{tab:acc_comp}
    \begin{tabular}{|c|c|}
        \hline
        Algorithm & Accuracy \\\hline
        {\em ExhaustiveSelection} & 0.7828 \\\hline
        {\em GreedySelection} & 0.7694 \\\hline
        {\em HybridSelection} (Our Approach) & {\bf 0.8075} \\\hline
    \end{tabular}
    
\end{table}

In Table \ref{tab:acc_comp}, we compare the recognition accuracy delivered by the three component selection algorithms. We limit the execution time of all algorithms to be 24 hours. The result suggests that {\em HybridSelection} produces the highest accuracy, which is even higher than {\em ExhaustiveSelection}. 
This is because {\em ExhaustiveSelection} can only complete enumerating up to 4 components within the time limit, which yields 78.28\% recognition accuracy at best. On the other hand, {\em GreedySelection} only inspects a very limited fraction of the search space. Therefore, its accuracy is the lowest.
This result demonstrates the advantage of {\em HybridSelection} - reach a balance between time efficiency and recognition accuracy. % 2
\section{Related Work}
\label{sc:related}
%We introduce the related work in two aspects: distance metric learning and handwritten Chinese character recognition.

\subsection{Distance Metric Learning}
% need
The need for appropriate ways to measure the distance or similarity between data points is almost ubiquitous in machine learning and data mining. This leads to the emergence of DML, which aims at automatically learning a suitable metric from data \cite{bellet2013survey}.
% linear v.s. non-linear
In terms of the format, there exist linear and non-linear metrics. Non-linear metrics, such as the two-histogram distance can capture non-linear variations in the data, however they give rise to non-convex formulations. Linear metrics, such as the Mahalanobis distance, are easier to optimize and thus attract much attention in metric learning. 

% Ensemble
 Chang et al. \cite{chang2012boosting} propose a boosting Mahalanobis distance metric (BoostMDM) method. It iteratively employs a base-learner to update a base matrix. A framework to combine base matrices is proposed in their paper, along with a base learner algorithm specific to it. The loss function describes the hypothesis margin, a lower bound of the sample margin used in methods such as SVMs (Support Vector Machines). 
% dropout
A problem with DML is that since the number of parameters to be determined is quadratic to the dimension of the features, it may lead to overfitting the training data \cite{weinberger2009distance}, and provide a suboptimal solution. 
To resolve the problem, Qian et al. \cite{qian2014distance} propose a regularization approach that applies dropout to both the learned metrics and the training data.

% MMC
The work quite closely related to ours is \cite{weinberger2009distance}. It aims at learning a Mahalanobis metric for clustering. The metric seeks to minimize the distance between similarly labeled inputs while maximizing the distance between differently labeled samples. However, this work does not specifically consider pictorial scripts with substantial diversity in OCR, nor does it focus on easily interpretable metrics. 
%AV: Boxiang, please modify this last sentence as needed to show how our work is different from theirs. 

\subsection{Handwritten Chinese Character Recognition}
Research on Chinese handwriting recognition of general texts has been observed only in recent years \cite{liu2007normalization}, and the reported accuracy is quite low. Recent works, using character classifiers and statistical language models (SLM) based on oversegmentation, obtain a character-level accuracy of $78.44 \%$ \cite{wang2009integrating} and $73.97 \%$ \cite{li2010bayesian}, respectively. Earlier works show even lower accuracy. 

Recent years witness the popularity of deep learning in fields such as natural language processing, computer vision and machine learning. Zhang et al. \cite{wu2017improving} obtain $95.88\%$ character-level accuracy on the CASIA-HWDB dataset by using a 15-layer convolutional neural network (CNN). Though accurate, the maxpooling and spatial pooling operations in the CNN render little interpretability in the recognition model.
Our work makes a contribution by providing good accuracy and easy interpretability, in addition to learning efficiency. 

 %1
\section{Conclusion}
\label{sc:conclu}
This paper addresses handwritten Chinese character recognition. We propose the {\em MetChar} algorithm  for distance metric learning, and the {\em HybridSelection} algorithm to select distance components.  By intelligently learning to combine base components, the learned distance metric has the desired interpretability, learning efficiency and recognition accuracy. Experiments on benchmark data reveal accuracy $\sim$ $81\%$. Further, we gain insights into the components most useful in Chinese character recognition, through this empirical study. 
%Experimental results on a benchmark dataset demonstrate the efficiency and precision of the proposed approach.
In the future, would investigate other optimization procedures to refine weight assignment for base components. We can consider the use of commonsense knowledge in learning ~\cite {Tandon2017} and investigate app development ~\cite{Basavaraju2016} based on related works. We plan to design classification techniques based on attentional neural networks. %0.25

\footnotesize

\section*{Acknowledgement}
This work incurs partial support through: startup funds for Dr. Boxiang Dong; a doctoral faculty program for Dr. Aparna Varde; and mentoring inputs from the NSF LSAMP grant for Danilo Stevanovic. %All these sources are from Montclair State University, NJ, USA. 

\bibliographystyle{IEEEtran}
{
\bibliography{bib}

\begin{thebibliography}{10}
\providecommand{\url}[1]{#1}
\csname url@samestyle\endcsname
\providecommand{\newblock}{\relax}
\providecommand{\bibinfo}[2]{#2}
\providecommand{\BIBentrySTDinterwordspacing}{\spaceskip=0pt\relax}
\providecommand{\BIBentryALTinterwordstretchfactor}{4}
\providecommand{\BIBentryALTinterwordspacing}{\spaceskip=\fontdimen2\font plus
\BIBentryALTinterwordstretchfactor\fontdimen3\font minus
  \fontdimen4\font\relax}
\providecommand{\BIBforeignlanguage}[2]{{%
\expandafter\ifx\csname l@#1\endcsname\relax
\typeout{** WARNING: IEEEtran.bst: No hyphenation pattern has been}%
\typeout{** loaded for the language `#1'. Using the pattern for}%
\typeout{** the default language instead.}%
\else
\language=\csname l@#1\endcsname
\fi
#2}}
\providecommand{\BIBdecl}{\relax}
\BIBdecl

\bibitem{apple}
``Real-time recognition of handwritten chinese characters spanning a large
  inventory of 30,000 characters.''
  \url{https://machinelearning.apple.com/2017/09/12/handwriting.html}, 2017.

\bibitem{chopra2005learning}
S.~Chopra, R.~Hadsell, Y.~LeCun \emph{et~al.}, ``Learning a similarity metric
  discriminatively, with application to face verification,'' in \emph{CVPR
  (1)}, 2005, pp. 539--546.

\bibitem{bellet2013survey}
A.~Bellet, A.~Habrard, and M.~Sebban, ``A survey on metric learning for feature
  vectors and structured data,'' \emph{arXiv:1306.6709}, 2013.

\bibitem{goldberger2004nips}
J.~Goldberger, S.~Roweis, G.~Hinton, and R.~Salakhutdinov, ``Neighborhood
  components analysis,'' in \emph{NIPS}, 2004, pp. 513--520.

\bibitem{yin2009handwritten}
F.~Yin and C.~L. Liu, ``Handwritten chinese text line segmentation by
  clustering with distance metric learning,'' \emph{Pattern Recognition},
  vol.~42, no.~12, pp. 3146--3157, 2009.

\bibitem{wang2011handwritten}
Q.~F. Wang, F.~Yin, and C.~L. Liu, ``Handwritten chinese text recognition by
  integrating multiple contexts,'' \emph{IEEE Trans. on Pattern Anal. and
  Machine Intelligence}, vol.~34, no.~8, pp. 1469--1481, 2011.

\bibitem{su2009off}
T.~H. Su, T.-W. Zhang, D.~J. Guan, and H.~J. Huang, ``Off-line recognition of
  realistic chinese handwriting using segmentation-free strategy,''
  \emph{Pattern Recognition}, vol.~42, no.~1, pp. 167--182, 2009.

\bibitem{wang2009integrating}
Q.~F. Wang, F.~Yin, and C.~L. Liu, ``Integrating language model in handwritten
  chinese text recognition,'' in \emph{IEEE Intl. Conf. on Document Anal. and
  Recognition}, 2009, pp. 1036--1040.

\bibitem{li2010bayesian}
N.~Li and L.~Jin, ``A bayesian-based probabilistic model for unconstrained
  handwritten offline chinese text line recognition,'' in \emph{IEEE Intl.
  Conf. on Systems, Man and Cybernetics}, 2010, pp. 3664--3668.

\bibitem{nishimura2003offline}
H.~Nishimura and T.~Timikawa, ``Offline character recognition using online
  character writing information,'' in \emph{IEEE Intl. Conf. on Document Anal.
  and Recognition}, 2003, pp. 168--172.

\bibitem{kingma2014adam}
D.~P. Kingma and J.~Ba, ``Adam: A method for stochastic optimization,''
  \emph{arXiv:1412.6980}, 2014.

\bibitem{wang2018ictai}
T.~Wang, J.~Huan, and B.~Li, ``Data dropout: Optimizing training data for
  convolutional neural networks,'' in \emph{IEEE ICTAI}, 2018, pp. 39--46.

\bibitem{varde2007icde}
A.~S. Varde, E.~A. Rundensteiner, G.~Javidi, E.~Sheybani, and J.~Liang,
  ``Learning the relative importance of features in image data,'' in \emph{IEEE
  ICDE DBrank workshop}, 2007, pp. 237--244.

\bibitem{varde2006sigmod}
A.~S. Varde, E.~A. Rundensteiner, C.~Ruiz, D.~Brown, M.~Maniruzzaman, and R.~D.
  Sisson~Jr, ``Effectiveness of domain-specific cluster representatives for
  graphical plots,'' in \emph{ACM SIGMOD IQIS workshop}, 2006, pp. 24--29.

\bibitem{vardedistance}
A.~Varde, S.~Bique, and D.~Brown, ``Distance metric learning by greedy,
  exhaustive and hybrid approaches,'' Tech. Rep, Virginia State University, VA,
  2007.

\bibitem{varde2008component}
A.~Varde, S.~Bique, E.~Rundensteiner, D.~Brown, J.~Liang, R.~Sisson,
  E.~Sheybani, and B.~Sayre, ``Component selection to optimize distance
  function learning in complex scientific data sets,'' in \emph{International
  Conference on Database and Expert Systems Applications}.\hskip 1em plus 0.5em
  minus 0.4em\relax Springer, 2008, pp. 269--282.

\bibitem{carbonetto2004statistical}
P.~Carbonetto, N.~De~Freitas, and K.~Barnard, ``A statistical model for general
  contextual object recognition,'' in \emph{European Conf. on Computer
  Vision}.\hskip 1em plus 0.5em minus 0.4em\relax Springer, 2004, pp. 350--362.

\bibitem{hossain2012rapid}
M.~Z. Hossain, M.~A. Amin, and H.~Yan, ``Rapid feature extraction for optical
  character recognition,'' \emph{arXiv:1206.0238}, 2012.

\bibitem{shen2012positive}
C.~Shen, J.~Kim, L.~Wang, and A.~Hengel, ``Positive semidefinite metric
  learning using boosting-like algorithms,'' \emph{JMLR}, vol.~13, no. Apr, pp.
  1007--1036, 2012.

\bibitem{qian2014distance}
Q.~Qian, J.~Hu, R.~Jin, J.~Pei, and S.~Zhu, ``Distance metric learning using
  dropout: a structured regularization approach,'' in \emph{ACM KDD}, 2014, pp.
  323--332.

\bibitem{chang2012boosting}
C.~C. Chang, ``A boosting approach for supervised mahalanobis distance metric
  learning,'' \emph{Pattern Recognition}, vol.~45, no.~2, pp. 844--862, 2012.

\bibitem{weinberger2009distance}
K.~Q. Weinberger and L.~K. Saul, ``Distance metric learning for large margin
  nearest neighbor classification,'' \emph{JMLR}, vol.~10, no. Feb, pp.
  207--244, 2009.

\bibitem{liu2007normalization}
C.~L. Liu, ``Normalization-cooperated gradient feature extraction for
  handwritten character recognition,'' \emph{IEEE Trans. on Pattern Anal. and
  Machine Intelligence}, vol.~29, no.~8, pp. 1465--1469, 2007.

\bibitem{wu2017improving}
Y.~C. Wu, F.~Yin, and C.~L. Liu, ``Improving handwritten chinese text
  recognition using neural network language models and convolutional neural
  network shape models,'' \emph{Pattern Recognition}, vol.~65, pp. 251--264,
  2017.

\bibitem{Tandon2017}
N.~Tandon, A.~Varde, and G.~{de Melo}, ``Commonsense knowledge in machine
  intelligence,'' \emph{ACM SIGMOD Record}, vol.~46, no.~4, pp. 49--52, 2017.

\bibitem{Basavaraju2016}
P.~Basavaraju and A.~Varde, ``Supervised learning techniques in mobile device
  apps for androids,'' \emph{ACM SIGKDD Explorations}, vol.~18, no.~2, pp.
  18--29.

\end{thebibliography}
}

\end{document}